\documentclass[a4paper, 10pt, conference]{ieeeconf}      

\IEEEoverridecommandlockouts                              
\usepackage[utf8]{inputenc}
\usepackage[T1]{fontenc}
\usepackage[table,xcdraw]{xcolor}

\usepackage{graphics} 
\usepackage{epsfig} 
\usepackage{hyperref}
\usepackage{multirow}
\usepackage{booktabs}
\usepackage{subfigure}

\title{\LARGE \bf
Museum Painting Retrieval
}

\author{{
\parbox{5 in}{
    \centering
    \`Oscar Lorente \qquad
    Ian Riera \qquad
    Shauryadeep Chaudhuri \\
    Oriol Catalan \qquad
    Víctor Casales\\
    Master in Computer Vision\\
    Universitat Aut\`onoma de Barcelona\\
    08193 Bellaterra, Barcelona, Spain\\
    {\tt\small\{oscar.lorentec, 
        ianpau.riera,
        shauryadeep.chaudhuri, \\
        oriol.catalan,
        victor.casales\}@e-campus.uab.cat
    }}
}}

\begin{document}

\maketitle
\thispagestyle{empty}
\pagestyle{empty}

\begin{abstract}
To retrieve images based on their content is one of the most studied topics in the field of computer vision. Nowadays, this problem can be addressed using modern techniques such as feature extraction using machine learning, but over the years different classical methods have been developed. In this paper, we implement a query by example retrieval system for finding paintings in a museum image collection using classic computer vision techniques. Specifically, we study the performance of the color, texture, text and feature descriptors in datasets with different perturbations in the images: noise, overlapping text boxes, color corruption and rotation. We evaluate each of the cases using the Mean Average Precision (MAP) metric, and we obtain results that vary between 0.5 and 1.0 depending on the problem conditions.
\end{abstract}


\section{INTRODUCTION}
\label{sec:intro} 
During the last decades, there has been a great increase in the number of digital image databases in applications such as medical imaging, geographic information or art collections, among others. For this reason, a method is needed to access, annotate, and catalog all this data efficiently. This topic has become a broad area of research interest, and one of the most studied methods are Content based image retrieval (CBIR) systems.

CBIR is a process in which for a query example image, similar images are obtained from a larger dataset based on the similarity in the content. There are different ways to extract content from an image, from descriptors as simple as color or illuminance histograms to key points and local descriptors. Today, advances in technology allow the use of advanced techniques such as machine learning to extract representative features from each image. However, to understand these concepts it is first necessary to understand traditional techniques.

In this paper, we present a query by example retrieval system for finding paintings in a museum dataset using various descriptors and datasets with different levels of difficulty. Regarding descriptors, we first explore color-based histograms by testing different color spaces (Gray, RGB) and types of histograms (1D, 3D, block or multiresolution). On the other hand, we study texture descriptors such as Histogram of Oriented Gradients (HOG), Discrete Cosinus Transform (DCT) or Local Binary Patterns (LBP). The names of the paintings authors are also used as text descriptors. Finally, we analyze the performance of the CBIR system using the Oriented FAST and Rotated BRIEF (ORB) feature detection algorithm \cite{orb}. Regarding the level of difficulty, there are datasets with noisy images, overlapping text boxes or more than one painting per image (which requires us to eliminate the background). The CBIR system architecture will need more complexity to deal with those cases.

\begin{figure}[t!]
    \centering
    \includegraphics[width=0.8\linewidth]{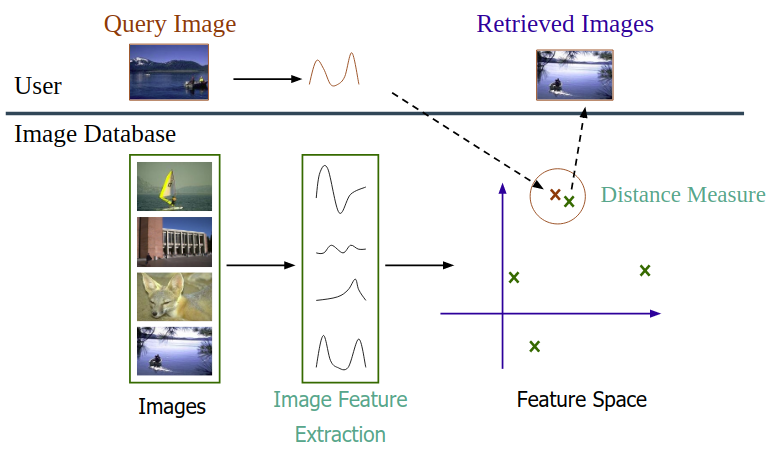}
    \caption{Simplified scheme of the CBIR system.} 
    \label{fig:cbir}
\end{figure}

\section{PROBLEM DEFINITION}
\label{sec:problem} 
Given a museum dataset and a query dataset, for each image in the query dataset, retrieve the K most similar images in the museum dataset, ordered by score. To do so, we first extract the features of the images from the museum dataset using a specific descriptor. Then, for each query image, we extract the corresponding features and compare them with those of the museum dataset images. This comparison is carried out using a specific distance as a metric (\textit{e.g.} Euclidean). Then, we order the museum images according to these distances. The museum image \textit{closest} to the query image is the retrieved image. In Figure~\ref{fig:cbir} we can see a simplified scheme of the system.

\section{DATA}
\label{sec:data}
The museum dataset contains 287 images, each of a different painting and with a unique label. The author and the name of each painting is also known.

\begin{figure}[t!]
\centering
\begin{tabular}{cccc}
\includegraphics[height=1.85cm]{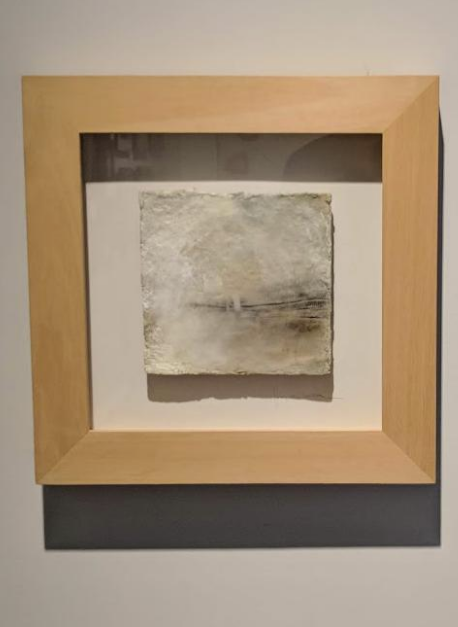} & 
\includegraphics[height=1.85cm]{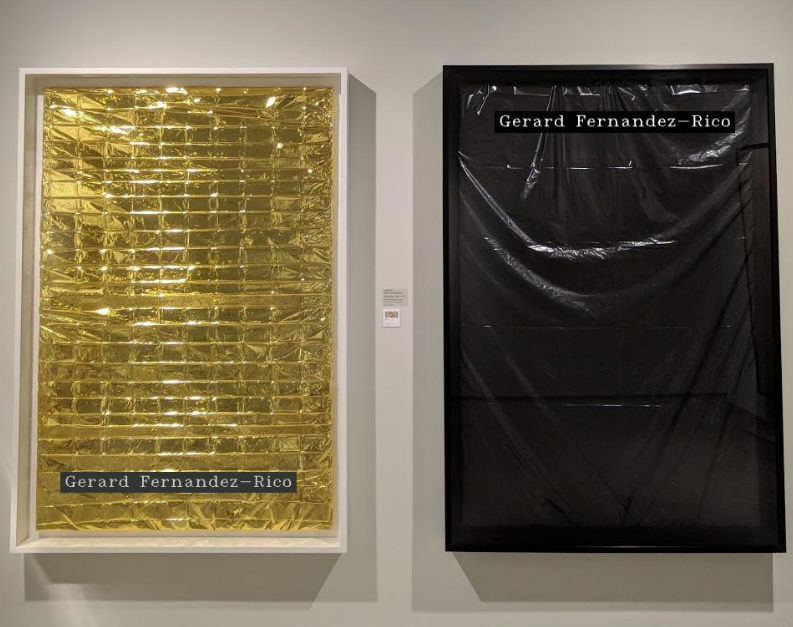} &
\includegraphics[height=1.85cm]{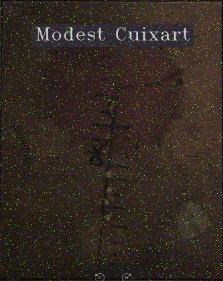} &
\includegraphics[height=1.85cm]{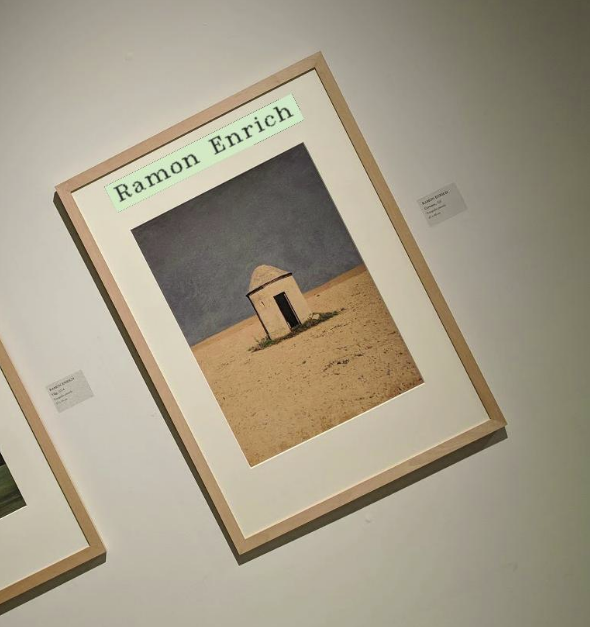} \\
(a) & (b) & (c) & (d)
\end{tabular}
\caption{Representative images from (a) Dataset 1 - Simple case; (b) Dataset 2 - Multiple paintings with text boxes; (c) Dataset 3 - Noisy images; (d) Dataset 4 - Rotated paintings} \label{fig:datasets}
\end{figure}

Regarding the query datasets, they all contain 30 images with up to three paintings per image and a background we must detect and remove. We work with four datasets of different levels of difficulty. In Figure~\ref{fig:datasets} we can see a representative image of each dataset.

\begin{itemize}
    \item \textbf{Dataset 1}: Only one painting per image. All the paintings exist in the museum database.
    
    \item \textbf{Dataset 2}: Up to two paintings per image. In addition, semi-transparent text boxes are superimposed to the paintings. The text boxes contain the name of the painter and have different sizes, fonts and locations.  All the paintings exist in the museum database.
    \item \textbf{Dataset 3}: Similar to dataset 2 but some images have noise and color corruption (random Hue changes). All the paintings exist in the museum database.
    
    \item \textbf{Dataset 4}: Similar to dataset 3 but some paintings are rotated, and not all of them are in the museum database.
    
\end{itemize}

\section{METHOD}
\label{sec:method}
In this section we explain in detail the methods used to implement a CBIR system in each of the datasets exposed in Section ~\ref{sec:data}. In each case, an attempt is made to minimize the impact of dataset disturbances and solutions and alternatives are presented to deal with them.
\subsection{Metrics}
To compare the query images with the database, different distance metrics were used:

\begin{itemize}
    \item Hellinger: 
    \begin{equation}\label{eq:hellinger}
    K\left(h_{1}, h_{2}\right)=\sum_{i=1}^{N} \sqrt{h_{1}(i) h_{2}(i)}
    \end{equation}
    
    \item Chi-squared:
    \begin{equation}\label{eq:chi-squared}
    D\left(h_{1}, h_{2}\right)=\sum_{i=1}^{N} \frac{\left(h_{1}(i)-h_{2}(i)\right)^{2}}{h_{1}(i)+h_{2}(i)}
    \end{equation}
    
    \item Intersection:
    \begin{equation}\label{eq:intersection}
    I\left(h_{1}, h_{2}\right)=\sum_{i}^{N} \min \left(h_{1}(i), h_{2}(i)\right)
    \end{equation}
    
    \item Correlation:
    \begin{equation}\label{eq:correlation}
    d\left(h_{1}, h_{2}\right)=\frac{\sum_{I}\left(h_{1}(I)-\bar{h}_{1}\right)\left(h_{2}(I)-\bar{h}_{2}\right)}{\sqrt{\sum_{I}\left(h_{1}(I)-\bar{h}_{1}\right)^{2} \sum_{I}\left(h_{2}(I)-\bar{h}_{2}\right)^{2}}}
    \end{equation} 
    where: 
    \begin{equation}\label{eq:histogram}
    \bar{h}_{k}=\frac{1}{N} \sum_{j}^{N} h_{k}(j)
    \end{equation}
    is the mean of the N histograms $h_{k}$.
\end{itemize}

Additionally, after retrieving the K most similar images from the museum database, the precision of the system is evaluated using the mAP@K (mean Average precision at K) metric of the complete query data set:
\begin{equation}\label{eq:mAP}
    AP@K = \frac{\sum_{i=1}^{K} P@i}{K}
\end{equation}
where \textit{P@i} is the precision obtained after retrieving \textit{i} images. In this paper, the mAP@1 is used to calculate the results.

Finally, to evaluate the method to detect text bounding boxes, we use the mean Intersection over Union (mIoU). The IoU is computed dividing the area of overlap between the predicted and groundtruth text boxes by its area of union.

\subsection{Dataset 1}
In the first dataset we have to solve two unknowns: what descriptors to obtain from the images and how to detect the background.

\begin{figure}[t!]
    \centering
    \includegraphics[width=0.65\linewidth]{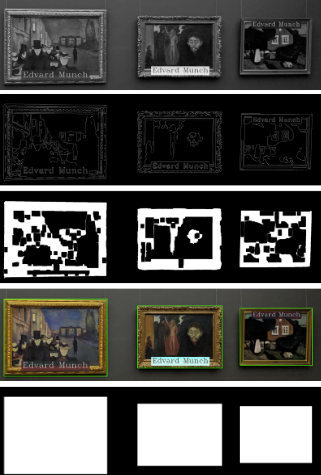}
    \caption{Background removal steps, from top to bottom: convert image to grayscale, detect edges with Canny, apply morphological closing, find contours and fill the masks.} 
    \label{fig:bg-removal}
\end{figure} 

\subsubsection{Background removal}
Before applying query retrieval we need to obtain the paintings (foreground) from the images. Various techniques can be used to detect and remove the background from an image. We have implemented a background detector using the painting contours and morphological operations, which is divided into the following steps:

\begin{enumerate}
    \item Convert the image to the gray color space
    \item Use Canny edge detector~\cite{canny}
    \item Apply morphological closing with a rectangle
    \item Find painting contours using ~\cite{contour-detector}
    \item Fill the mask
\end{enumerate}

In this way, we detect the background accurately and we can extract the painting of interest from the image. These steps are visually represented in Figure~\ref{fig:bg-removal}. To evaluate the performance of the background removal system, we calculate the precision, recall and F$_{1}$-score comparing the results to the groundtruth masks.

\subsubsection{Color Descriptors}
The information of an image can be represented by color histograms. Using different color spaces and dimensionalities of the histogram, we can generate a great variety of descriptors of the same image. In this paper, we have tested the system using 1D histograms in the grayscale and 3D histograms in the RGB, LAB, HSV and YCrCb color spaces. The results are presented in Table \ref{tab:color-space-distance-map}.

\begin{figure}[t!]
    \centering
    \includegraphics[width=0.8\linewidth]{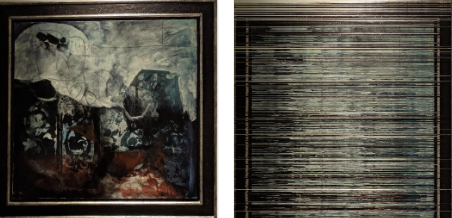}
    \caption{Two different images with the same histogram.} 
    \label{fig:problem-histogram}
\end{figure}

\begin{table}[t!]
\centering
\caption{mAP comparison between descriptors using different color spaces and metrics on dataset 1.}
\label{tab:color-space-distance-map}
\begin{tabular}{c|cccc}
\toprule
\multirow{2}{*}{\textbf{Color Space}} & \multicolumn{4}{c}{Distance Metric} \\
{} & Correlation & $Chi^{2}$ & Intersection & \textbf{Hellinger} \\ 
\midrule
\midrule
Gray & 0.23 & 0.22 & 0.09 & 0.26 \\
\textbf{RGB} & 0.33 & 0.16 & 0.21 & \textbf{0.53} \\
LAB & 0.39 & 0.22 & 0.23 & 0.49 \\
YCrCb & 0.34 & 0.21 & 0.19 & 0.47 \\
HSV & 0.32 & 0.19 & 0.15 & 0.45 \\
\bottomrule
\end{tabular}
\end{table}

The 3D RGB histograms using the Hellinger distance (see Equation~\ref{eq:hellinger}) is the configuration with the best mAP results. Note that we use 3D histograms instead of concatenating the 1D histograms of each of the 3 channels, as it represents the number of pixels that have the same R, G and B values at the same time. 

Despite the results, this configuration is not robust, since the lack of spatial information can lead to paintings erroneously paired, as shown in Figure~\ref{fig:problem-histogram}.

To better preserve spatial information and thus improve the system we use block-based histograms. This method consists of dividing the image into blocks and obtaining the color histogram of each of the blocks, which results in a more representative descriptor.

\begin{figure}[t!]
    \centering
    \includegraphics[width=0.8\linewidth]{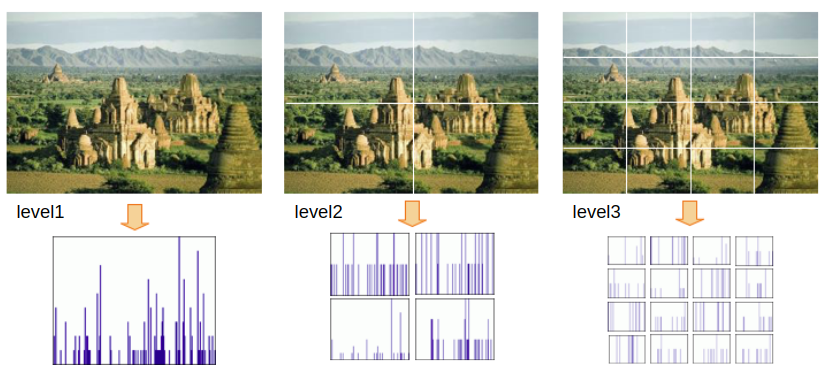}
    \caption{Representation of a multiresolution histogram, which concatenates block-based histograms at different levels.} 
    \label{fig:multires-histograms}
\end{figure}

Another alternative is the multiresolution histograms, which allow obtaining a spatial pyramid representation of the image. Specifically, the block-based histograms method is applied at different levels, to later concatenate the histogram at each level (Figure~\ref{fig:multires-histograms}).

The results of each method are presented in Table \ref{tab:color-descriptors-map}. We use 16x16 blocks for the block-based method and 4 levels (1x1, 4x4, 8x8 and 16x16) for the multiresolution histogram. The best results are obtained with the block-based descriptor. For this reason, when the \textbf{color descriptor} is mentioned in this paper, it will be referring to \textbf{16x16 blocks-based histogram}.

\begin{table}[t!]
\centering
\caption{mAP comparison between color descriptors (histograms methods) on dataset 1.}
\label{tab:color-descriptors-map}
\begin{tabular}{cc}
\toprule
Color Descriptor & mAP \\ 
\midrule
\midrule
3D RGB & 0.53 \\
Block-based & 0.83 \\
Multiresolution & 0.8 \\
\bottomrule
\end{tabular}
\end{table}

\subsection{Dataset 2}
In the second dataset, the background removal method must be used to extract up to two paintings per image. Moreover, we also need to detect and remove the superimposed text bounding boxes to ensure that they do not affect the result.

\subsubsection{Text bounding box detection}
The overlapping text bounding boxes are semi-transparent and have different colors in each image. To detect them, we propose a robust method that works on all three channels of the LAB color space.

First, for each channel (\textit{L}, \textit{A} and \textit{B}) we apply the morphological operators of \textbf{top hat} and \textbf{black hat} separately.

\begin{itemize}
    \item Top hat: difference between the image and the \textit{opening} of the image. The opening removes the small bright elements, so the result of top hat highlights these small bright elements (\textit{e.g.} bright letters).
    
    \item Black hat: difference between the \textit{closing} of the image and the image. Completely opposite, the closing eliminates the small dark elements, so the result of black hat highlights these small dark elements (\textit{e.g.} dark letters).
\end{itemize}

\begin{figure*}[h!]
    \centering
    \begin{tabular}{cc}

    \includegraphics[width=0.55\linewidth]{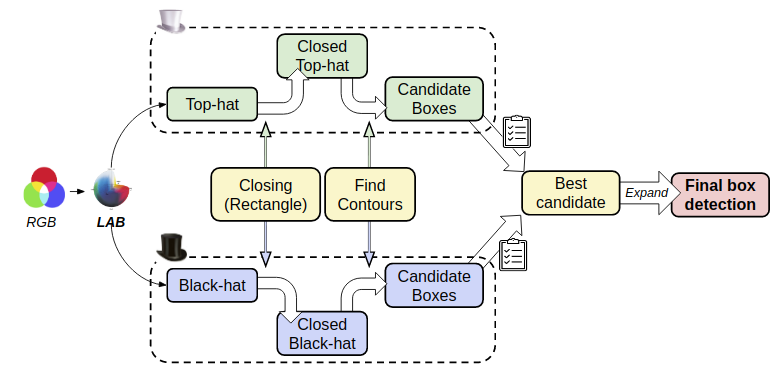} &
    \includegraphics[width=0.15\linewidth]{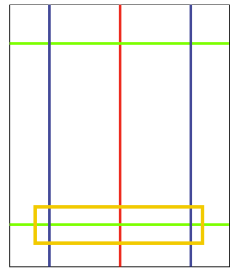} \\
    (a) & (b)
    
    \end{tabular}

    \caption{(a) Text boxes detection steps, from left to right: convert image to LAB color space, apply top hat and black hat for each channel, apply morphological closing, find contours and return the best box candidate using the distance score system; (b)  Text boxes scores: X center point (red line), Y center point (green lines) and symmetry of the box with respect to 1/5 and 5/6 of the image width (blue lines).} 
    \label{fig:detect-boxes}
\end{figure*}

Whether the letters of the text are dark or light, when applying the top hat and black hat operations separately, they are highlighted and thus easier to detect. The next step is to apply a closing operator using a rectangle to connect the letters and then apply \cite{contour-detector} to detect the contours. This process is shown in Figure~\ref{fig:detect-boxes}a.

With the presented method, both top hat and black hat operations are applied separately, so we have more than one candidate to be the text bounding box. To decide whether to keep the detection of the top hat or that of the black hat, we define a scoring system that compares all the candidates and returns the best one (Figure~\ref{fig:detect-boxes}b).

We use the following indicators to calculate the distance score:

\begin{itemize}
    \item X Center point: Distance between the bounding box center and the x center of the image
    
    \item Y Center point: The boxes are usually at the bottom or top of the image, so we take the 1/5 and 4/5 of the image height as reference points. 
    
    \item Symmetry of the box with respect to the center x line.

    \item Symmetry of the box with respect to 1/5 and 5/6 of the image width. The boxes borders are usually at the 1/6 and 5/6 of the image width, so we take them as reference points.
    
    \item Aspect ratio of the bounding box: $width \approx 4 * height$
\end{itemize}

Each of this indicators is weighted and added. The bounding box with the minimum resulting distance is returned as the detected box. 

By applying this method for each LAB channel, we can obtain up to three final candidates to be the detected text bounding box. Again, the scoring system is used to get the best of the three detections. To evaluate the performance of our method, we use the mean intersection over union to compare the predicted text box with the groundtruth.

\subsection{Dataset 3}
In the third dataset, we have to remove the noise added to some random images. Moreover, since there are images with color corruption, we cannot depend on the color descriptors anymore, so we need to define some new descriptors for the CBIR system. 

\subsubsection{Noise removal}
Noise affects only certain random images in the dataset, so we have two unknowns to solve: how to detect which images are noisy and how to remove their noise.

\begin{figure}[t!]
\centering
\begin{tabular}{cc}
\includegraphics[width=0.35\linewidth]{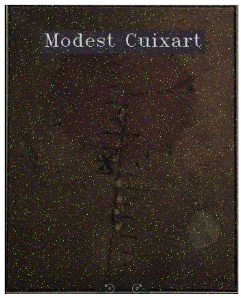} &
\includegraphics[width=0.35\linewidth]{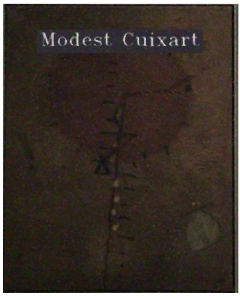} \\
(a) & (b)
\end{tabular}
\caption{(a) Noisy image; (b) Denoised image} \label{fig:noise-removal}
\end{figure}

For each image, we apply a median filter and check if the Peak Signal-to-Noise Ratio (PSNR) of the resulting image with respect to the original one is higher than a threshold. If that is the case, then the median filter is improving the image quality, so the image contains noise that needs to be removed (Figure~\ref{fig:noise-removal}). Otherwise, if the PSNR is lower than the threshold, the original image is preserved.

\subsubsection{Texture descriptors}
The corruption on the colors of some images cannot be corrected, so we need descriptors not related to color. Our first approach is to use the following texture descriptors:

\begin{itemize}
    \item \textbf{Local Binary Patterns (LBP)}: For each pixel, LBP compares its value to the values of its surrounding neighbouring pixels and labels the pixel with a decimal number based on this comparison. By concatenating the labels of each pixel, we obtain a feature vector with which to compare the paintings.
    
    \item \textbf{Discrete Cosine Transform (DCT)}: It expresses an image in terms of a sum of cosine functions oscillating at different frequencies, which helps us to obtain representative coefficients of the image.
    
    \item \textbf{Histogram of Oriented Gradients (HOG)}: This technique divides an image in portions (\textit{i.e.} blocks) and calculates the gradient and orientation for each portion to further count gradients occurrences.
\end{itemize}

\begin{figure*}[t!]
    \centering
    \includegraphics[width=0.8\linewidth]{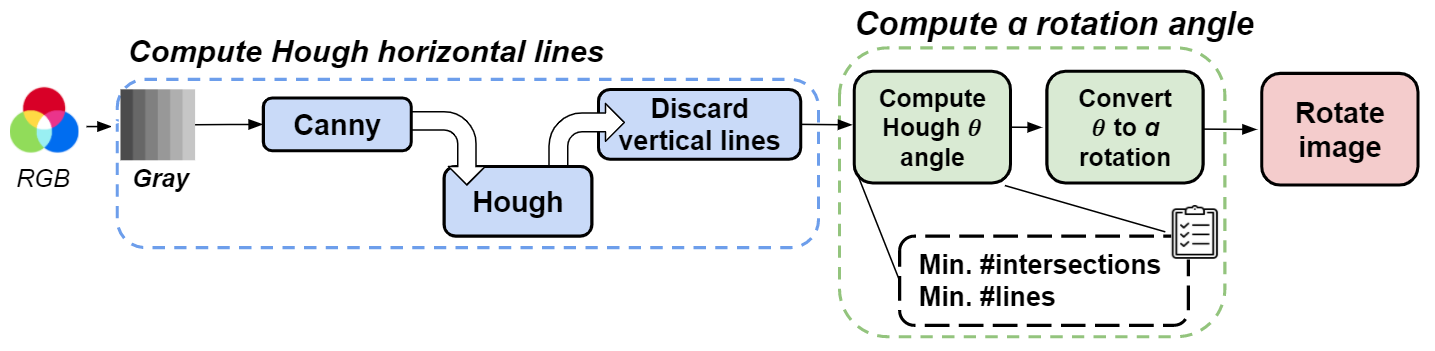}
    \caption{Rotation correction algorithm using Hough lines transform.} 
    \label{fig:hough}
\end{figure*}

\begin{table}[t!]
\centering
\caption{Comparison between texture descriptors on dataset 3.}
\label{tab:texture-descriptors-map}
\begin{tabular}{cc}
\toprule
Texture Descriptor & mAP \\ 
\midrule
\midrule
LBP & 0.44 \\
DCT & 0.49 \\
HoG & 0.72 \\
\bottomrule
\end{tabular}
\end{table}

The results are presented in Table \ref{tab:texture-descriptors-map}. The HOG descriptor is the one that provides better results. For this reason, when the \textbf{texture descriptor} is mentioned in this paper, it will be referring to \textbf{HoG}.

\subsubsection{Text descriptors}
Taking advantage of the detections of the text bounding boxes, an alternative to the texture descriptors is to detect the text inside the bounding boxes, which contains the name of the painter. In this way, knowing \textit{a priori} the author of each painting in the museum dataset, the names detected in the query image can be used to obtain the corresponding museum image.

To extract the text from the bounding box, we first binarize the content and then apply Optical Character Recognition (OCR) techniques \cite{ocr} from the Pytesseract library. Then, the resulting text is compared with the painters names in the database to see which one is more similar. 

The problem with this descriptor is that there might be more than one painting with the same author name in the museum's database. Therefore, with the text descriptors we cannot directly obtain the painting of interest, but we can use them as complementary help to the other descriptors.

\subsubsection{Combining descriptors}
Another approach is to combine all the aforementioned descriptors: color, texture and text. When combining descriptors, we first compute the distances for each descriptor (\textit{e.g.} Hellinger distance for color descriptor). Then, these distances are weighted and added to obtain a final distance result. The image with the smaller distance is the retrieved most similar painting.

\subsection{Dataset 4}
Finally, the fourth dataset presents two additional problems: some paintings are rotated and not all of them are in the museum database.

\subsubsection{Rotation of the images}
In order to apply some of the preprocessing methods presented in this paper (\textit{e.g.} text box detection), we need to find the rotation angle of each image and revert the rotation. To do so, we design two different methods. 

For the first one, we apply Canny edge detection and morphological closing to detect the paintings contours. Then, we find the rotated rectangles of the minimum area enclosing the contours of interest. From the two lower coordinates of each painting, we calculate the slope of the line that connects them and obtain the angle of rotation as the arc-tangent of the slope.

The second method is based on the Hough transform. In the gray color space, we apply Canny to detect the edges, and then use Hough to obtain the larger lines. As we are interested in the rotation with respect to the x axis, the vertical lines are discarded. After removing outliers, we compute the mean rotation angle using the remaining horizontal lines. This process is shown in Figure~\ref{fig:hough}.

In both cases, after obtaining the rotation angle, we revert the rotations and then apply the preprocessing methods: background removal, denoising and text box removal.

\subsubsection{Feature descriptors}
In all the previous descriptors, we calculate the distances between a query image and all the database images and retrieve the image with the minimum distance, \textit{closest} to the query one. Therefore, an image is always retrieved even if the query image and the database one are completely different. To be able to compare images properly and detect paintings non existent in the museum dataset, the features of the images (corners, gradients, key points) are extracted and compared. Two images are considered similar if a minimum number of features matches is achieved. If this minimum threshold is never exceeded, the query image is considered non existent in the museum database.

\begin{figure}[t!]
\centering
\subfigure[]{\includegraphics[width=0.4\textwidth]{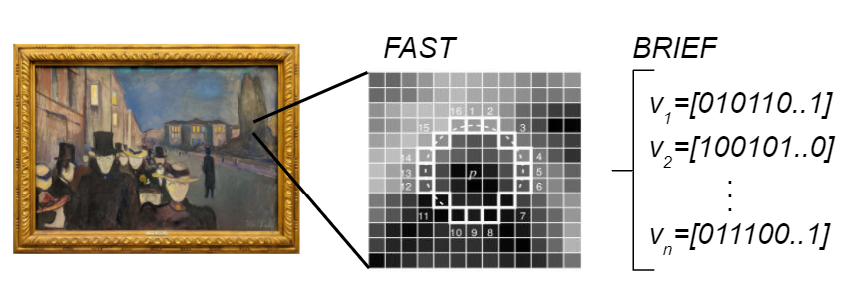}}
\subfigure[]{\includegraphics[width=0.3\textwidth]{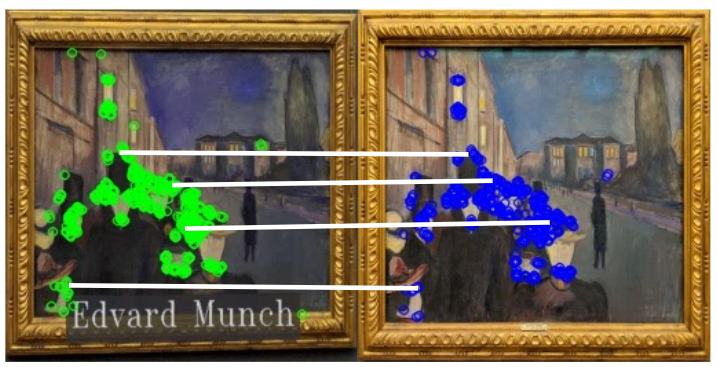}} 
\caption{ORB method: (a) FAST to obtain keypoints and BRIEF to vectorize the features; (b) Painting matching example} \label{fig:ORB}
\end{figure}

In this paper, the \textbf{Oriented FAST and Rotated BRIEF (ORB)} feature detection algorithm \cite{orb} is used. This method (Figure~\ref{fig:ORB}a) relies on the FAST keypoint extraction, which given a pixel p compares the brightness of p to the surrounding 16 pixels in a circle. Pixels in the circle are then sorted into three classes: lighter than p, darker than p or similar to p. If more than 8 pixels are darker or brighter than p, then it is selected as a keypoint.

Keypoints found by FAST give us information of the location of determining edges in an image. BRIEF takes all these keypoints and convert them into a binary feature vector so that together they can represent an object. The paintings keypoints are compared and considered as a match if the distance between them is smaller than a threshold. If there are 4 or more matches between two images (Figure~\ref{fig:ORB}b), then they are considered similar. The most similar image (\textit{i.e.} with more matches) is retrieved as the corresponding museum image.

The ORB algorithm is invariant to rotation, so there is no need to revert the rotation of the paintings. However, it is not invariant to noise, background and superimposed text boxes. For example, if the text boxes are not removed, the algorithm would detect unwanted keypoints on the text. Therefore, the system performs better with the complete preprocessing pipeline.

\section{RESULTS}
\label{sec:results}

\begin{table}[t!]
\centering
\caption{Evaluation of the background removal method on each dataset.}
\label{tab:background-evaluation}
\begin{tabular}{{c|cccc}}
\toprule
Dataset & Precision & Recall & F$_{1}$-score & mAE \\ 
\midrule
\midrule
1 & 0.90 & 1.0 & 0.94 & -  \\
2 & 0.91 & 0.99 & 0.94 & - \\
3 & 0.91 & 1.0 & 0.95 & - \\
4 & 0.90 & 0.99 & 0.94 & 0.43º \\
\bottomrule
\end{tabular}
\end{table}

Regarding the background detector method, in Table~\ref{tab:background-evaluation} we present the results obtained when comparing the detected backgrounds with the corresponding groundtruth. As can be seen, we obtain a full recall in the four datasets, which means that all the paintings are correctly detected. In global terms, the implemented background removal method is robust and performs satisfactorily. 

For the last dataset, the mean Angular Error (mAE) between the estimated angles of rotation of the paintings and the groundtruth is presented. The results are satisfactory, since an error of only 0.43º allows us to confirm that the method has been implemented correctly.

On the other hand, we have evaluated the performance of the descriptors presented in this paper for each of the four datasets. Specifically, we present the results obtained in datasets 1, 2, 3 and 4 using descriptors of color, texture, text, a combination of these and finally features. A mAP@1 is used at all times.

The results are presented in Table \ref{tab:evaluation-descriptors}. As can be seen, the color descriptors work fairly well except in the datasets where there are paintings with color corruption (datasets 3 and 4). In these cases, the texture descriptors provide better results. On the other hand, using only text descriptors we cannot detect the paintings of interest properly, as there are ambiguous cases in which different paintings are from the same painter. Combining the three descriptors, the performance of the system improves remarkably in each of the cases. However, it continues to fail in dataset 4, since we cannot identify the paintings that are not in the museum dataset with only this information. This problem is solved using the feature descriptors, which not only allow detecting these cases, but also improve the results in each one of the datasets.

\begin{table}[t!]
\centering
\caption{Evaluation of the descriptors on each dataset.}
\label{tab:evaluation-descriptors}

\begin{tabular}{c|cccc}
\toprule
\multirow{2}{*}{\textbf{Descriptor}} & \multicolumn{4}{c}{Dataset} \\
{} & 1 & 2 & 3 & 4 \\ 
\midrule
\midrule
Color & 0.83 & 0.7 & 0.64 & 0.42\\
Texture & 0.93 & 0.89 & 0.72 & 0.53 \\
Text & - & 0.21 & 0.16 & 0.13 \\
Combined & 0.97 & 0.86 & 0.78 & 0.46 \\
\textbf{Feature} & \textbf{0.97} & \textbf{0.9} & \textbf{0.92} & \textbf{0.98} \\
\bottomrule
\end{tabular}
\end{table}

\section{CONCLUSIONS}
\label{sec:conclusions}
In this paper, we have explored some traditional techniques that can be used in a CBIR system, and we have seen the advantages and disadvantages of using some descriptors compared to others. From the results obtained, we can draw the following conclusions. 

First of all, color descriptors are not efficient in cases where images present color corruption. In those cases, texture descriptors provide better results, as they are invariant to the color corruption. Additionally, in datasets 1 and 2, texture descriptors also improve the performance of the system with respect to the color ones, so we can conclude that texture can better represent the information of the image.

Secondly, in our specific case, text descriptors are not enough to retrieve the paintings correctly, as there might be more than one painting per author. However, it can be combined with the color and texture descriptors to improve the overall system performance.

Finally, feature descriptors are the ones that provide the best results, since they represent the information of the images  accurately, at the cost of using more complex and time consuming algorithms.

In terms of background detection, the overall results are satisfactory. The precision is high but could we improved, as with the current method the shades under the paintings frames are wrongly detected.

We have also implemented methods for solving problems such as noisy images, superimposed text bounding boxes or image rotation. By designing and using the proper preprocessing algorithms, the performance of the resulting CBIR system is greatly improved.

\section{IMAGE CLUSTERING}
Using the descriptors explained in this paper, we present an online museum exhibition by clustering the paintings of the museum dataset. We use the K-means algorithm to cluster the images into 10 different groups. K-means analyzes the data to find organically similar data points and assigns each point to a cluster with similar characteristics. Each cluster can then be used to label the data into different classes based on its characteristics.

The images are first clustered by colour into two sets of clusters: bright and dark images. Then, both sets are clustered by texture (using HOG descriptor) into 5 more clusters, resulting in 10 different clusters. For each of the bright and dark clusters, with the texture clustering we obtain a range from plain to highly detailed images. The resulting 10 clusters are labeled conveying a distinct mood. Feel free to visit our online exhibition \cite{museum-exhibition}.

\bibliographystyle{unsrt}
\bibliography{bibliography}
\addtolength{\textheight}{-12cm}   

\end{document}